# Étude et traitement automatique de l'anglais du XVII<sup>e</sup> siècle : outils morphosyntaxiques et dictionnaires

*Hélène Pignot et Odile Piton*

Laboratoire SAMM-Marin Mersenne – Université Paris1 Panthéon-Sorbonne

RÉSUMÉ.
Après avoir exposé la constitution du corpus, nous recensons les principales différences ou particularités linguistiques de la langue anglaise du XVII<sup>e</sup> siècle, les analysons du point de vue morphologique et syntaxique et proposons des équivalents en anglais contemporain (AC). Nous montrons comment nous pouvons effectuer une transcription automatique de textes anglais du XVII<sup>e</sup> siècle en anglais moderne, en combinant l'utilisation de dictionnaires électroniques avec des règles de transcriptions implémentées sous forme de transducteurs.

ABSTRACT.
In this article, we record the main linguistic differences or singularities of 17[th] century English, analyse them morphologically and syntactically and propose equivalent forms in contemporary English. We show how 17[th] century texts may be transcribed into modern English, combining the use of electronic dictionaries with rules of transcription implemented as transducers.

MOTS-CLÉS: anglais du XVII<sup>e</sup> siècle, étude diachronique, traitement dynamique, dictionnaires électroniques, analyse morphologique, graphes morphologiques, grammaires syntaxiques, transducteurs.

KEYWORDS: 17[th] century English, diachronic study, dynamic processing, electronic dictionaries, morphological analysis, morphological graphs, syntactic grammars, transducers.

**Introduction**

L'anglais du XVII<sup>e</sup> siècle présente de nombreuses particularités orthographiques, syntaxiques et lexicales qui en font tout le charme mais également la difficulté. Même pour un locuteur natif, la lecture de ces beaux textes qui font partie du patrimoine littéraire et historique de l'humanité, n'est pas chose aisée. Les ouvrages du XVII<sup>e</sup> siècle réédités pour les étudiants ou pour le grand public sont donc assortis d'un riche apparat critique, et de multiples notes concernant le lexique et éclairant les circonstances historiques de leur rédaction.

Notre domaine de recherche conjugue deux passions, pour l'histoire anglaise et européenne au XVII<sup>e</sup> siècle et pour le récit de voyage. Six années durant nous avons pu redécouvrir les érudits et voyageurs français et anglais en Grèce et en Anatolie (Pignot 2007, 2009). En traduisant les textes anglais pour l'édition française ou en les établissant pour l'édition anglaise, nous avons pu mesurer leur difficulté et leur richesse. Une question a jailli dans notre esprit : comment rendre ces textes plus accessibles aux lecteurs modernes non spécialistes et aux étudiants, comment éviter qu'ils ne soient rebutés par une langue qu'ils jugeront peut-être archaïque et difficile à comprendre ?

Nous avons donc souhaité œuvrer au développement d'un outil automatique permettant d'appréhender la langue de cette époque. A partir de notre corpus nous recensons les différences ou particularités linguistiques de la langue de cette époque, les analysons du point de vue morphologique et syntaxique et proposons des équivalents en anglais contemporain (AC). Notre collaboration, qui a débuté en 2006, visait d'abord à définir des outils permettant de formaliser les données linguistiques nécessaires à la traduction diachronique automatique (dictionnaires et grammaires bilingues). Depuis, nous avons construit, appliqué, testé et affiné ces outils, et présentons ici nos résultats.

Le présent article comportera trois volets, l'un présentant le corpus, l'autre consacré à l'analyse morphologique et syntaxique automatique de l'anglais du XVII<sup>e</sup> siècle, et le dernier à l'étude du lexique, en particulier à la constitution d'un dictionnaire électronique recensant les particularités lexicales de l'anglais du XVII<sup>e</sup> siècle (mots archaïques ou dont le sens a évolué en anglais moderne) et proposant – quand cela est possible – un équivalent en AC. Notre objectif ultime est de créer un dictionnaire de l'anglais du XVII<sup>e</sup> siècle, ainsi que des outils facilitant la lecture des récits de voyage du XVII$^e$ siècle quelle que soit leur aire géographique, et leur accès, par exemple pour le grand public ou pour un public d'étudiants (les implications pédagogiques seront évoquées dans la conclusion).

**1. Constitution et exploitation du corpus**

Pour les chercheurs, de nombreux textes du du XVII$^e$ siècle sont accessibles grâce à la base de données EEBO (*Early English Books Online*). Les textes sont ainsi téléchargeables au format 'pdf' (image). Au moment où nous avons constitué le corpus, nous n'avions pas accès à EEBO ni à des logiciels d'OCR. À l'époque de la sélection de ces textes nous avons fait le choix de retaper tous les extraits relatifs à la Grèce.

Constatant une évolution des fontes d'imprimerie, nous avons dû opérer quelques modifications. Au cours de notre opération de saisie nous avons remplacé le « s long » par des s et remplacé l'éperluette par « and » (on ne peut donc rechercher les occurrences de ces caractères dans notre corpus). Il faut noter que la différentiation entre i et j, u et v n'est effective qu'à partir de 1634, comme le remarque Robert Lass (Hogg, Lass et al. 1992). On recommande aux imprimeurs d'éviter certains doublements de consonnes et l'ajout de e muet final mais les pratiques varient, voir les **Images 1, 2 et 3**.

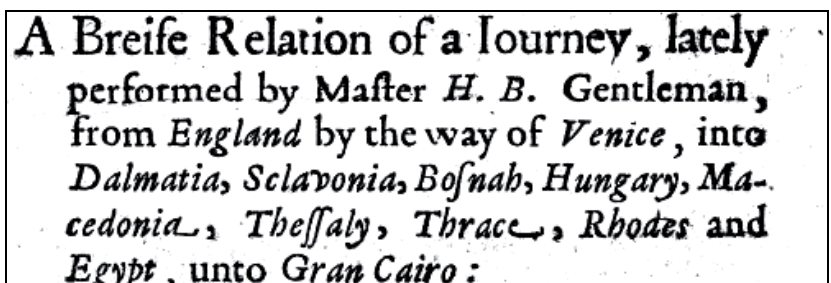

**Image 1.** Exemple de typographie de l'anglais du XVII$^e$.

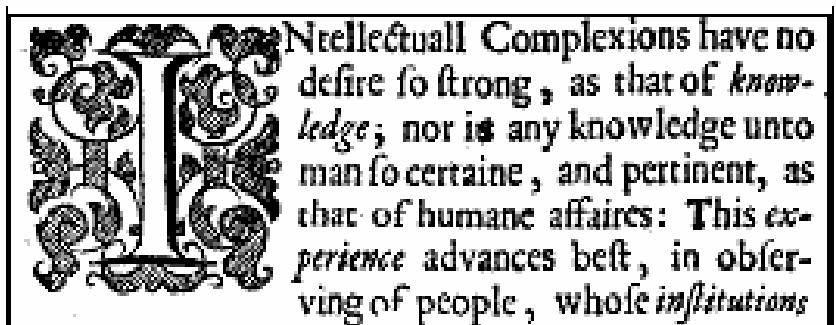

**Image 2.** Exemple de typographie de l'anglais du XVII$^e$.

| XVII$^e$ | AC | XVII$^e$ | AC | XVII$^e$ | AC |
|---|---|---|---|---|---|
| &c | etc. | feuerall | several | foueraignrie | sovereignty |
| vnder | under | *Mofes.* | Moses | feafon | season |
| *Iewes* | Jews | haue | have | flauery | slavery |
| approoued | approved | ioyneth | joins | difpoffeffeth | dispossesses |

**Image 3.** Typographie de l'anglais du XVII$^e$.

Les auteurs qui constituent notre corpus sont George Sandys, Henry Blount, John Ray, Paul Rycaut, Thomas Smith et George Wheler. Chronologiquement le premier est le poète George Sandys (1578-1644), fils d'un ecclésiastique, Edwin Sandys, qui fut archevêque d'York. C'est peut-être pour se consoler d'un mariage malheureux et de ses diverses complications que Sandys décida de s'embarquer pour le Levant. Selon l'historien Hugh Trevor Roper, George Sandys désirait l'union des églises et portait de ce fait un certain intérêt à la situation de l'Église grecque dans l'Empire ottoman. L'itinéraire emprunté par notre voyageur lui permet de visiter la France, le Nord de l'Italie, la Turquie (qu'il explore une année durant), l'Égypte, la Palestine, Chypre, la Sicile, Naples, et Rome. Sa relation de voyage, qui parut sous le titre, *A Voyage to the Levant*, fit florès, connaissant sept rééditions au cours du XVII[e] siècle. Il y consacre quelques pages aux coutumes des Grecs, à leur foi, à leur mode de vie, à leurs coutumes, et à leur langue.

Le second auteur est Henry Blount (1602-1682), avocat de profession, qui s'embarqua à Venise sur un navire en direction du Levant, en compagnie de voyageurs turcs et juifs. Désireux de découvrir la Turquie et de comprendre les raisons de sa puissance, il en profita pour faire connaissance des peuples qui vivent sous le joug ottoman, « les Grecs, les Arméniens, les Francs, les Tsiganes, et particulièrement les Juifs ». Sa relation de voyage, *A Voyage into the Levant*, fut publiée pour la première fois en 1636, et connut huit éditions entre 1636 et 1671. Ce qui retient l'attention de ce voyageur plus sceptique de nature, ce sont surtout les renégats du christianisme.

John Ray (1627-1705) était pour sa part botaniste. Il n'a pas visité l'Empire ottoman mais a parcouru l'Europe trois années durant, découvrant la Hollande, l'Allemagne, la Suisse, l'Italie, la Sicile et Malte. Dans son ouvrage *A Collection of Curious Travels and Voyages through the Levant* (1693), Ray réunit les contributions de divers voyageurs dans l'Empire ottoman et résume les chapitres que Pierre Belon, botaniste français et voyageur au Levant au milieu de XVI[e] siècle consacre au Mont Athos dans un de ses récits de voyage.

Quant à Paul Rycaut (1629-1700), il fit carrière dans la diplomatie d'abord comme secrétaire de l'ambassadeur du roi d'Angleterre Charles II à Constantinople auprès du Sultan Mahomet IV, puis de 1667 à 1678, comme consul d'Angleterre à Smyrne. Cette position d'observateur privilégié lui permettra de publier en 1679 *The Present State of the Greek et Armenian Churches*, qui n'est pas seulement un témoignage sur ces deux Églises et sur leurs coutumes, mais également une relation de voyage puisque le texte comprend la description de certains lieux visités par l'auteur tels l'Asie Mineure et le Mont Athos. Rycaut souhaitait également l'union des Églises et rédigea ce livre sur les instances du roi Charles II, qui désirait parfaire sa connaissance de la foi des orthodoxes grecs.

Thomas Smith (1638-1710) étudia les langues orientales et fut *fellow* de Madgalene College à Oxford. En qualité de chapelain, il accompagna Sir Daniel Harvey, ambassadeur d'Angleterre à Constantinople et demeura trois années à son service (de 1668 à 1671). En 1672 il visita les sept Églises d'Asie et en donna une description dans un ouvrage rédigé en latin et publié en 1694. Dans sa correspondance, lui aussi s'interroge sur la possibilité d'une union entre l'Église d'Angleterre et celle de Grèce, et il publie en 1680 un ouvrage intitulé *An Account of the Greek Church*, où il expose les rites et les doctrines de cette Église.

Pour clore cette liste, le botaniste anglais George Wheler (1650-1723), partit à la découverte de la Grèce en compagnie du médecin et archéologue lyonnais Jacob Spon (1647-1685). Ils publièrent chacun un récit de leur voyage (Wheler, 1682). À son retour de Grèce, Wheler fut ordonné pasteur. Il fut l'un des premiers membres des *Society for the Propagation of the Gospel* et *Society for Promoting Christian Knowledge*, et la Grèce lui semblera un terrain propice à l'activité missionnaire anglicane.

Ainsi, ce qui rapproche nos auteurs, qui voyagent à des moments historiques très différents, c'est la curiosité pour la Grèce et le désir de comprendre ce qui différencie les Grecs des autres chrétiens. Ces différences seraient-elles de nature à empêcher une union avec l'Église d'Angleterre ? Rycaut et Smith se posent clairement la question. Toutes ces précisions historiques sont indispensables pour nous permettre de comprendre pourquoi un certain nombre des exemples de notre troisième partie sera emprunté au vocabulaire religieux : la société grecque dépeinte par nos voyageurs est organisée autour de ses croyances religieuses, la vie est rythmée par le calendrier ecclésiastique, les fêtes religieuses, les carêmes, les prières et tous les rites de l'Église.

Ces textes nous fournissent un échantillon linguistique qui balaie tout le siècle (de 1615 à 1693). Ce corpus est multilingue. Il comporte de nombreuses citations en grec ancien, en latin, des emprunts plus ou moins nombreux selon les langues, au turc, au grec, à l'italien et à l'espagnol. Il nous permet de dresser un inventaire des différences orthographiques, lexicales et syntaxiques entre l'anglais du XVIIe siècle et l'anglais moderne ; on remarquera une certaine cohérence.

**2. Étude diachronique : modélisation des évolutions de la langue**

Toponymes et ethnonymes étaient souvent différents en anglais du XVIIe siècle, ainsi que leur orthographe, ce qui engendre un double dépaysement pour le lecteur, plongé dans un lointain univers culturel, d'ailleurs très souvent qualifié d'« oriental » par nos voyageurs lors même qu'il s'agit des Grecs, et doté d'une toponymie différente et pour le moins déroutante parfois ! Donnons quelques exemples : Nice est un autre nom de Nicée en Asie Mineure, Romagnia désigne la Grèce et les Balkans, le Maina est le sud du Péloponnèse (Mani en anglais moderne), la Croatie s'appelle Sclavonia, les îles de la mer Egée situées entre la Grèce et la Turquie ou îles de l'Archipel se nomment aussi « the Arches ».

Le vocable « Hollanders » désigne les Hollandais, « Grecians » les Grecs (et pas seulement ceux de l'Antiquité comme en anglais moderne), les Zynganaes ou « Zinganies », les tziganes.

Parmi les termes historiquement datés, nous avons également fait le relevé de toutes les unités de poids et mesure, de distance spécifiques au XVIIe siècle ainsi que des monnaies, qui ne sont guère parlantes pour un lecteur non averti. Nous avons aussi répertorié les titres étrangers. Voir la Table1.

| Catégorie | Exemple | Définition |
|---|---|---|
| Monnaie | *A zecchine* | Sequin, pièce vénitienne |
|  | *A hungar* | Pièce hongroise |
|  | *A sultany* | Pièce turque en or |
|  | *An asper* | Pièce turque en argent |
|  | *A ducat* | Pièce italienne en or ou en argent utilisée dans toute l'Europe |
| Unité de poids | *An oque* | 1kg 250 |
| Unité de distance | *A league* | circa 4 km |
|  | *A furlong* | 200 m |
| Titre ou profession turc | *Bassa* | le Pacha |
|  | *A cadi* | un juge |
|  | *A dragoman* | un interprète |
|  | *Keslar-Agafi* | Eunuque noir qui surveille les femmes du Sérail |
| Titre grec | *Egoumen* | le Père Abbé d'un monastère |
| Titre italien | *the Grand Signior* | le Sultan |
|  | *the Bailo* | le Baile vénitien |

**Table 1.** Termes historiques: exemples.

Nous avons fait le recensement de tous les mots étrangers et proposé un équivalent en anglais moderne. Les emprunts ont une fonction d'authentification du témoignage et manifestent le degré d'implication du voyageur dans la société où il vit. En recourant à ces mots empruntés au grec et au turc, nos auteurs veulent montrer la spécificité des us et des coutumes décrits (il y a de très nombreux termes grecs et citations en grec dans les textes de Rycaut et de Smith en particulier). La Table 2 récapitule quelques-uns de ces emprunts.

| Langue | Emprunts lexicaux | Anglais contemporain |
|---|---|---|
| Arabe | *salam'd;'salamed'* | saluted |
| Français | *randevouzes* | pluriel du mot français rendez-vous, utilisé pour le rendez-vous galant en AC |

| Grec | *antidoron* | blessed bread |
|---|---|---|
| | *comparos* | godfather |
| | *diataxis* | order |
| | *douleia* | veneration of a saint or of relics and icons |
| | *Eikonomachoi, or Eikonoklastai* | those who oppose the worship of images or destroy them |
| | *kaloir (kalogieros)* | a good elder: a Greek Orthodox monk |
| | *kosmokratores* | Emperors of the world |
| | *latinophrones* | latinizing |
| | *latreia* | worship of God |
| | *mamoukode* | a ghost |
| | *metanoïa* | repentance |
| | *metousiosis* | modification |
| | *Paranomoi* | flagitious persons, and transgressors of the laws and canons of the Church |
| | *somatikos* | corporal |
| | *sponsalia* | the betrothal ring |
| | *ta trimera* | third day after death, on which prayers are said for the departed soul |
| | *tèn diairesin apo tês alethe* | a disunion from the truth |
| | *tó déma* | "tying up a man from accompanying with any woman" a spell to make a man impotent |
| | *to katasphragisai to paidion* | the sealing of infants |
| | *vroukolakas* | an evil spirit |
| Italien | *canaglia* | a rascal |
| | *a capriccio of the Grand Signior* | a caprice of the Sultan |
| | *Madonna di Constantinopoli* | the blessed Virgin of Constantinople (an icon believed to have been painted by St. Luke) |
| Latin | *dissenteaneous* | contrary (latin *dissentaneus*) |
| | *flagitious* | guilty of terrible crimes (latin *flagitiosus*, OED[1] 1550) |
| | *supposititious* | spurious (latin *supposititius*, OED, 1611) |
| | *margaritae* | pearls or particles of the Holy Communion |
| Turc | *alempena* | Constantinople or the refuge of the world |
| | *baratz* | Commission from the Grand Signor |
| | *bezesten* | market-place |
| | *gazi* | conqueror |
| | *harach* | poll-money (tribut) |
| | *kabin* | cohabitation as opposed to marriage |
| | *kara congia* | a demon appearing in the shape of a black old man |

**Table 2.** Exemples de mots empruntés.

Enfin la langue du XVIIe siècle comporte un certain nombre d'archaïsmes, de formes anglicisées de mots latins et grecs totalement inusités en anglais moderne, et de mots dont le sens a évolué. La Table 3. fournit quelques exemples significatifs, les catégories sont codées N pour les noms, A pour les adjectifs, et V pour les verbes. En faire le relevé et proposer un équivalent en anglais moderne facilite la compréhension de ces textes non seulement par des lecteurs dont la langue maternelle n'est pas l'anglais mais aussi par des Anglophones. Toutes ces particularités sont répertoriées dans notre dictionnaire.

| Anglais du XVIIe siècle | Catégorie | Anglais contemporain |
|---|---|---|
| Alcoran | N | the Koran |

| arbitrement | N | Arbitration |
| --- | --- | --- |
| chane (Arabic, khan) | N | an inn |
| constitute | V | to set up in an office or position of authority |
| declension | N | a decline |
| dissentaneous | A | contrary to |
| drubbing | N | Beating |
| ethnick (E) | A | pagan, heathen (grec, ethnos, OED 1470) |
| gossip | N | a godfather or godmother |
| grogoran | N | grogram, a coarse fabric of silk, mohair and wool |
| penitentiary | N | a spiritual father ou a penitent (selon le contexte) |
| pix | N | a vessel in which the consecrated bread of the Sacrament is kept |
| prejudicacy | N | prejudice |
| runnugate | N | a renegade |
| shash | N | a sash, a scarf worn around the waist |
| symbolize with someone | V | to resemble, to partake of the nature of |
| tenent | N | a tenet |
| turcism | N | Islam |
| upstart (E) | A | lately come into existence |

**Table 3.** Archaïsmes ou termes dont le sens a évolué (E).

## 3. Méthodologie : traitement morphologique et syntaxique

Nous avons opté pour l'utilisation de la plate-forme linguistique NooJ. Rappelons brièvement qu'elle permet de construire des outils sous forme de dictionnaires de formes fléchies – ces formes étant toutes associées au lemme correspondant (verrai, voyais, vis sont associées au lemme voir) – ou sous forme de grammaires exprimées par des règles morphologiques ou syntaxiques. Elle effectue à la demande l'analyse lexicale d'un texte par des outils sélectionnés (dictionnaires et grammaires), ce qui l'indexe, produit en retour des informations sur le texte, notamment la liste des formes reconnues, et celle des formes non reconnues. De nombreux travaux ayant déjà porté sur la langue anglaise contemporaine, des dictionnaires et des graphes sont mis à disposition (Silberztein 1993, 2003). *A contrario*, les particularités de la langue du XVII$^e$ ne sont pas prises en charge. C'est soit au moyen de dictionnaires, soit au moyen de grammaires que nous allons définir les outils qui vont permettre de pré-analyser automatiquement un texte du XVII$^e$, ainsi que les propositions de réécriture de mots ou fragments de textes en AC. Nous obtenons ainsi des entrées lemmatisées qui, après validation, vont enrichir notre dictionnaire. Notre but est double : constituer un dictionnaire et affiner nos outils de reconnaissance et de traitement.

Les entrées d'un dictionnaire électronique sont de la forme : lemme,code catégorie+trait1+ …+traitn. Les codes de catégories sont respectivement N, A, V, PRO… pour les noms, les adjectifs, les verbes et les pronoms…. Les traits syntaxiques sont insérés lors de la compilation du dictionnaire, selon le modèle indiqué sous la forme FLX=codeflexion. Le trait +EN=xxx permet d'indiquer la transcription en AC du mot concerné. Les différents codes sont indiqués dans la documentation NooJ, qui est accessible en ligne à l'adresse suivante <www.nooj4nlp.fr>. Nous verrons que nous pouvons être conduits à établir des ordres de priorité entre certains mots, ce qui revient à constituer *plusieurs dictionnaires* et à les hiérarchiser selon un système de priorité.

Insistons sur le fait que l'approche du traitement informatique est bien différente de l'approche linguistique, en conséquence des traitements automatiques similaires peuvent concerner des aspects linguistiques très différents, et inversement des variations d'un même élément linguistique peuvent nécessiter des outils de traitement de nature totalement différente. Nous présentons une étude détaillée, et en indiquons le mode opératoire : création de nouvelles formes lemmatisées, et règles de réécriture des mots isolés et des syntagmes identifiés, que nous pouvons automatiser. Enfin nous proposerons un bilan de notre travail et des difficultés rencontrées.

Il importe de faire clairement la distinction opérée par NooJ entre grammaire morphologique et grammaire syntaxique : une *grammaire morphologique*, permet de traiter un lexème unique, tandis qu'une *grammaire syntaxique* traite de formes disjointes, lesquelles peuvent concerner des lexèmes distincts, ou appartenir au même lexème comme lors de l'utilisation de /'d/ pour remplacer une flexion ed (*dry'd* pour dried). Nous les présenterons sous forme de graphes.

L'observation des transformations, et leur aspect rare ou systématique donne lieu à deux traitements qui sont complémentaires : une modification isolée est traitée par une entrée dans le dictionnaire électronique, qui comportera la réécriture du mot en AC, tandis qu'une modification qui nous paraît avoir un caractère aspect répétitif, est écrite sous forme de règle permettant d'identifier des lexèmes dans le texte, et de produire automatiquement des propositions d'entrées pour le dictionnaire. Nous verrons l'importance de l'ordre d'application des règles ainsi décrites.

L'examen de l'anglais du XVII$^e$ nous permet d'observer des différences orthographiques telles que le redoublement de consonnes, des différences de préfixes et de suffixes, l'ajout ou la suppression de e muets. Certaines lettres ont deux formes : u et w, i et j comme en latin classique. Certains mots sont dissociés en deux lexèmes alors qu'en anglais moderne ils sont concaténés, comme "for ever" ou "sun-set", ou inversement comme dans le mot monyworth, ancienne forme de l'expression "money's worth". La ponctuation est différente : les deux-points sont utilisés soit pour marquer la fin d'une phrase, soit pour marquer une pause dans la phrase. L'éperluette (&) remplace souvent la conjonction "et".

Nous présentons en premier lieu le traitement de formes grammaticales, puis nous traiterons des variations morphologiques et des expressions disjointes.

### 3.1. Les modifications grammaticales

–Les noms et pronoms : nombre de ces mots comportent deux composants séparés, parfois reliés par un trait d'union, alors qu'en AC ils sont concaténés (par exemple any thing, any one, church-yard, ou Arch-Angel) ou au contraire, North wind est écrit Northwind. La transformation de deux lexèmes reliés par un trait d'union en lexème unique se fait par un simple graphe qui concatène les deux éléments. La Figure 2 présente le graphe qui permet de transcrire automatiquement mid-land en <midland> ; countrey-men en <countrymen>.

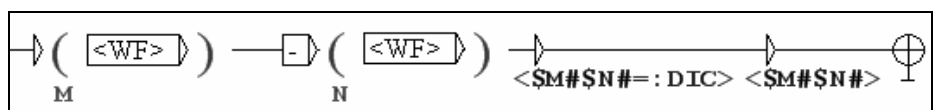

**Figure 2.** Suppression du trait d'union.

Inversement la transformation d'un lexème unique en mot composé ou en expression disjointe se fait au moyen d'entrées dans le dictionnaire. Soient monyworth,N+EN="money's worth" et Northwind,N+s+EN="North wind".

–Le génitif, qui s'écrivait sans apostrophe : "from the *womens* apartment"; "out of their wives and childrens mouths" commence à s'orthographier /'s/; mais outre le génitif —dans "the grand Signior's women"—, on rencontre la flexion /'s/: pour marquer le pluriel interlingual de noms étrangers, comme dans *Egoumeno*'s, *Bassa*'s (Pashas), ou *piazza*'s qui sont reconnaissables par le graphe présenté en Figure 3.

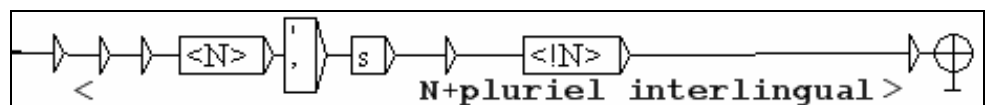

**Figure 3.** Pluriel interlingual.

Le premier graphe présenté en Figure 4 est un graphe morphologique qui permet de marquer un mot inconnu terminé par un s, comme anothers (dans "one anothers company"), ou childrens (dans "their wives and childrens mouths"). Remarquons que wives est également un génitif pluriel, il est analysé comme un pluriel et la conjonction and permettra sa reconnaissance par le graphe syntaxique. Celui-ci identifie la succession <N+gen_sax> <N>, ou <PRO+gen_sax> <N>, soit childrens mouths et anothers company. Il propose les transcriptions : <another's company> et < wives's and children's mouths>[2].

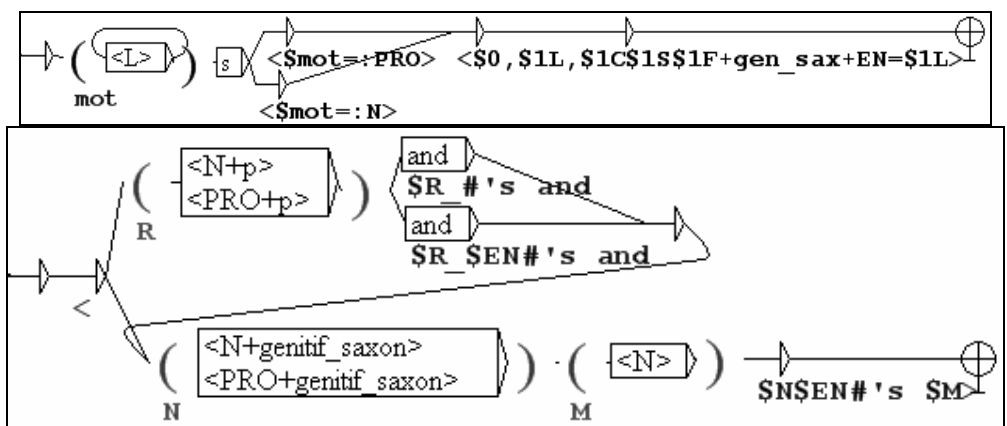

**Figure 4.** Identification du génitif saxon et insertion de /'s/.

Notons que mans dans « no other mans errors could draw » n'est pas identifié comme un génitif, car il est faussement reconnu comme le verbe man.
– Pronoms et adjectifs.
–Les pronoms personnels peuvent s'orthographier hee, shee, wee, ye, thay au lieu de he, she, we, you, et they. Ceci se traite simplement par le dictionnaire : hee,PRO+m+s+EN=he ; shee,PRO+m+s+EN=she etc. La 2e personne du singulier –thou, thee, thine = tu, toi, tien– est souvent utilisée (particulièrement dans les prières) : "We offer to thee thine of thine own"[3], in all things and through all things" (Smith).
–Les pronoms réflexifs ne sont pas lexicalisés, les deux lexèmes sont encore séparés : it self(e), my self(e), him self(e) ), your selves "Take heed unto your selves". Le dictionnaire permet d'enregistrer des mots composés. La correspondance entre les deux formes se fait donc simplement : it self,PRO+EN=itself. On peut le fléchir pour créer les deux formes équivalentes it self et it selfe. Ces entrées sont affectées du trait +UNAMB qui les désambiguïse.
–Les pronoms relatifs. Though peut s'orthographier tho'. Le pronom "which" peut être utilisé lorsque l'antécédent est humain, ce qui n'est plus le cas en AC: "The Georgians, which in some manner depend on the Greek Church, baptize not their children until they be eight years of age" (Rycaut).
–Morphologie verbale et paradigmes verbaux
L'adaptation des outils de flexion est requise, car les formes verbales correspondantes ne sont pas reconnues par les outils existants pour l'AC : quand il s'agit de flexion archaïque d'un lemme (verbe, pronom…) existant en AC, il faut construire un outil permettant d'ajouter la flexion manquante.
–Au présent, deux formes sont remarquables: la 2e personne du singulier, qui reçoit le morphème de flexion /est/ comme dans "thou satisfyest", et la 3e personne du singulier qui prend le morphème flexionnel /(e)th/, comme dans "shee cometh", "he hath", "he doth". Ces formes pouvant être rencontrées pour n'importe quel verbe, il faut les traiter dynamiquement par un graphe morphologique.

```
adviseth,advise,V+Tense=PR+Pers=3+Nb=s+EN=advise
desireth,desire,V+Tense=PR+Pers=3+Nb=s+EN=desire
saith,say,V+Tense=PR+Pers=3+Nb=s+EN=say
establisht,establish,V+Tense=PP+ EN=establish
establisht,establish,V+Tense=PT+Pers=1+Nb=p +EN=establish
fixt,fix,V+Tense=PP +EN=fix
fixt,fix,V+Tense=PT+Pers=1+Nb=p +EN=fix
```

> linkt,link,V+Tense=PP +EN=link

**Table 4.** Entrées ajoutées au dictionnaire pour le présent et le prétérit.

–Au prétérit, certains verbes, qui sont irréguliers en AC, ont une forme fléchie en /ed/ (ex. : catched est le prétérit de catch, shined est le prétérit de shine), d'autres verbes dont l'infinitif se termine en p, k, x, ss, sh ont deux prétérits possibles, normal ou raccourci, ed devenant t ou 't. D'autres verbes ont des formes différentes, par exemple to speak (I spake) or to begin (I begun, au lieu de began en AC). Ces formes seront insérées dans le dictionnaire.

L'apostrophe peut être utilisé pour le prétérit, le participe passé –voir la Figure 5–, ou l'adjectif déverbal qui prend alors la forme xyz't ou xyz'd. Deux situations peuvent se présenter : xyz, forme tronquée, est néanmoins « reconnu » à tort (cas de establish'd), ou peut ne pas l'être (cas de judg'd). C'est un graphe syntaxique comportant les séquences <WF>'t ou <WF>'d qui complètera la reconnaissance. Notre corpus comporte 90 occurrences de 'd ou 't.

Established se rencontre sous les trois formes : "and establish'd the same number", "the establish't doctrine", "and for ever established the adoration". Nous opérons la reconnaissance de ces formes au moyen d'un graphe morphologique, puis d'un graphe syntaxique, présentés Figure 5. Establish est faussement reconnu comme l'infinitif ou le présent. Le graphe syntaxique permet de construire established, qui est soit le participe passé, soit le prétérit, soit l'adjectif déverbal associés au verbe establish.

Étudions les cas de judg'd : "this being judg'd", la forme « judg » n'est pas identifiée par les dictionnaires. Nous appliquons les graphes morphologiques chargés de tester si moyennant certaines modifications, nous pouvons faire une hypothèse pour la reconnaissance de cette forme. La flexion ed, postposée à "judg" permet de construire judged le participe passé de judge. Le graphe syntaxique va confirmer cette interprétation et nous obtenons l'interprétation : judg,judge,PV+Tense=PP+ EN=judge.

Le mot professed est présent sous trois formes qui sont : professed, profes'd et profest : although "professed enemies to the Roman Church", "At the time of their being profest", "they believe Christianity can hardly be professed". Le cas de profest a été traité dans la Table 4. Il faut identifier « profes ». Le graphe ôtant un e muet émet la proposition profes,prof,N+Nb=p+Distribution=Hum+EN=prof, soit le pluriel de prof. Un autre graphe ajoute sed, et propose d'identifier professed, soit : profes,profess,PV+Tense=PP+EN=profess, (le graphe complet propose aussi de reconnaître le prétérit et l'adjectif déverbal). Par suite, le graphe syntaxique vérifie la présence de 'd et cela invalide la première possibilité, et valide la deuxième.

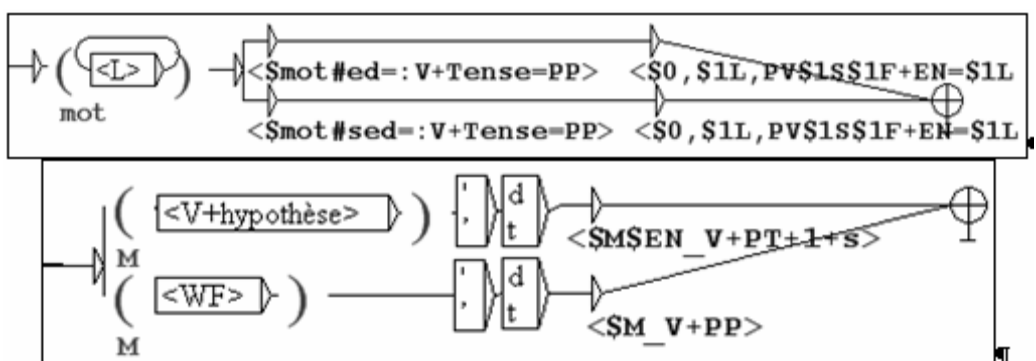

**Figure 5** Graphes permettant d'identifier le participe passé en 'd.

Le cas de imbrac'd est différent car pour reconnaître embraced il faut effectuer les transformations : imbrac'd → embrac'd → embraced. Il faut donc combiner deux transformations.

–Prépositions et conjonctions. Though peut s'orthographier tho'. Through est parfois écrit thro' (e.g. "thrusting an iron stake thro' the body out under the neck"). Les prépositions sont souvent omises : "but our subject here is more tragical, the subversion of the sanctuaries of religion, the royal priesthood expelled their churches, and those converted into mosques" (ici from est omis).

–Déterminants : leur utilisation est très proche de celle de l'AC excepté pour l'article indéfini "an"

devant des mots commençant par h comme dans "an holy amulet". On trouve aussi bien "an horrid impiety" que "a horrid sound". La transformation an → a est assez aisée car en AC on recense peu de mots commençant par un h non aspiré.
–"Wh-compounds" : les formes whosoever, wheresoever et whensoever sont couramment utilisées. Ces trois mots sont à ajouter au dictionnaire sous la forme: whosoever,PRO+EN=whoever.

### 3.2. Variations morphologiques et mots non reconnus

Il faut souligner l'aspect multilingue et religieux lié au contexte historique, géographique et culturel de ces textes, qui sont des récits de voyage, et à la propension des auteurs à recourir aux emprunts lexicaux que nous devons identifier et insérer avec les traits convenables à notre dictionnaire électronique. Notre corpus comporte des termes provenant de l'arabe, du grec, du latin, de l'italien et du turc.

### 3.2.1. Archaïsmes, variations sémantiques et entités nommées

L'indexation d'un lexème du XVII$^e$ siècle par les lemmes de l'AC peut conduire à différentes situations :
–Variation sémantique : les lexèmes sont reconnus, mais le sens a changé, ce sont les variations les moins flagrantes car le dictionnaire électronique pourra indexer faussement le mot qui ne sera pas repérable à ce niveau. Seul le lecteur averti peut repérer ce phénomène. Nous devons constituer un dictionnaire de tels termes et le rendre plus prioritaire que le dictionnaire anglais de NooJ. La reconnaissance prioritaire de penitentiary avec le sens de "spiritual father" va empêcher la comparaison de ce lexème aux dictionnaires moins prioritaires. Nous indiquons en Table 5, quelques entrées du dictionnaire, le code de flexion FLX=Nsp_y permettra de construire automatiquement le pluriel en remplaçant le y final par ies, soit penitentiaries.

```
penitentiary,N+FLX=Nsp_y+EN="spiritual father"+XVII
penitentiary,N+FLX=Nsp_y+EN="penitent"+XVII
ethnick,A+EN=pagan+ XVII
```
**Table 5**. Exemples d'entrées de dictionnaire prioritaires.

La priorité entre dictionnaires induit non seulement un ordre d'application, mais aussi des interdictions : les dictionnaires de niveau inférieur ne sont appliqués que si tous les dictionnaires de niveau de priorité supérieur n'ont pas produit de résultat.
–Le mot n'existe plus en anglais moderne, il est archaïque, ou bien le lexème étudié est la concaténation de plusieurs lexèmes en AC, il faut l'ajouter au dictionnaire.
–Le mot n'existe pas en anglais moderne parce qu'il s'agit d'une abréviation ou encore d'une Entité Nommée (EN) – par exemple d'un toponyme – or les ressources dictionnairiques ne comportent pas ou peu d'EN. Il faut les insérer dans le dictionnaire. Voir la Table 6.

```
Augustus,N+PR+m+s+Hum
Bajazid,N+PR+m+s+Hum+EN=Bajazet
Basileis Romaion,N+PR+p+Hum+EN="Emperor of the Greeks"
Nice,N+PR+s+Toponyme+EN=Nicæa
Romagnia,N+PR+s+Toponyme+EN="Greece and the Balkans"
```
**Table 6**.Dictionaire de toponymes et de mots empruntés.

### 3.2.2. Variations morphologiques

Le mot peut avoir subi une transformation morphologique comme le redoublement de lettres, l'insertion/suppression du e muet, ou une modification comme le remplacement de /edge/ par /ege/. Ce sont des transformations morphologiques dont nous pouvons décrire le paradigme, et traiter ces cas par un transducteur qui identifie, teste et traite le lexème, construit le lemme moderne correspondant et propose une entrée pour le dictionnaire.

Certaines des modifications peuvent se retrouver aussi bien à l'intérieur du mot qu'au début ou à la fin. Dans ce cas, dans le paradigme, soit X, soit Y est vide. Nous présentons quelques exemples en Table 7, et la Figure 6 présente des graphes de transformation..

| Modification | Exemples | Paradigme |
|---|---|---|
| im pour em | imbrace pour embrace | XimY → XemY |
| en pour in | encreasing pour increasing | XenY → XinY |
| in pour en | Intangling, intrench pour entangling, entrench | XinY → XenY |
| ncy pour nce | occurrency pour occurrence | XncyY → XnceY |
| ous pour ate | degenerous pour degenerate | XousY → XateY |
| ick pour ic | Arabick, garlick pour Arabic, garlic | XickY → XicY |
| ik pour ic | traffik pour traffic | XikY → XicY |
| ie pour y | christianitie, pour christianity | XieY → XyY |
| ie, ey pour y | countrey pour country | XeyY → XyY |
| our pour or | emperour, terrour pour emperor, terror | XourY → XorY |
| oa pour o | shoar, cloaths, choake | XoaY → XoY |
| edge pour ege | alledge, colledge pour allege, college | XedgeY → XegeY |
| ai pour ei | soveraigne | XaiY → XeiY |
| ea pour e | compleate, seaven | XeaY → XeY |
| ph pour f | phantastique | XphY → XfY |
| w pour u | perswasion pour persuasion | XwY → XuY |
| y pour i | oyl, coyn pour oil, coin | XyY → XiY |
| i pour j | iourney pour journey, Iew pour Jew | XiY → XjY |
| eer pour ear | yeer, neer pour year, near | XeerY → XearY |
| ence pour ense | expence | XenceY → XenseY |

**Table 7.** Exemples de modifications morphologiques.

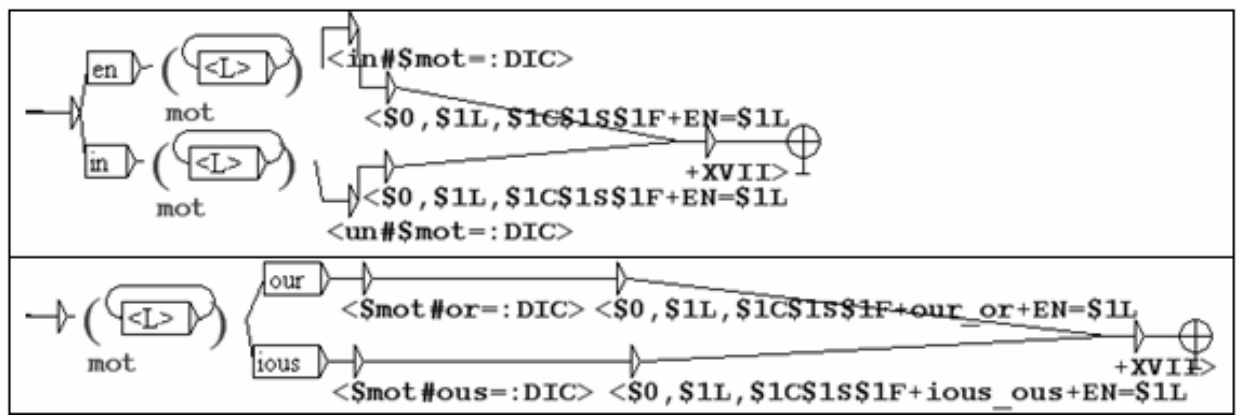

**Figure 6.** Graphes de transformations morphologiques.

L'application de ces graphes permet d'identifier encreasing comme l'adjectif et le nom increasing, ou le verbe increase, et governours comme le nom pluriel governor. La Table 8 présente des entrées du dictionnaire. Le trait +XVII marque la diachronie. Le mot identifié comme étant en anglais du XVII$^e$ est suivi de sa forme moderne, de sa catégorie codée et de traits syntaxiques, le trait EN=XXX indique que l'équivalent en AC est XXX.

```
encreasing,increasing,A+EN=increasing+XVII
encreasing,increase,N+Nb=s+EN=increase+XVII
encreasing,increase,V+Tense=G+EN=increase+XVII
```

> governours,governor,N+Nb=p+Distribution=Hum+EN=governor+XVII
> inferiour,inferior,N+Nb=s+Distribution=Hum+EN=inferior+XVII
> inferiour,inferior,A+EN=inferior+XVII

**Table 8.** Entrées proposées par le graphe.

### 3.3. Traitement de formes discontinues : reconnaissance et règles de réécriture

–Le subjonctif (qui est formé grâce au verbe à l'infinitif sans to à toutes les personnes) est très commun. Il apparaît dans les clauses conditionnelles et concessives (contrairement à l'AC où il est utilisé principalement dans des contextes formels). Il est précédé de unless | lest | if | provided | though | whether | whatever etc., comme dans "provided it be done after a due manner", "if they be kaloirs" (Smith), "whatever it be" (Smith), mais aussi dans des clauses temporelles, "until they be eight years of age" (Rycaut), et dans les clauses comparatives, "they had rather remaine as they be" (Sandys). Le graphe présenté en Figure 7 réécrit "provided it be" en "provided it is" et "if the criminal be" en "if the criminal is".

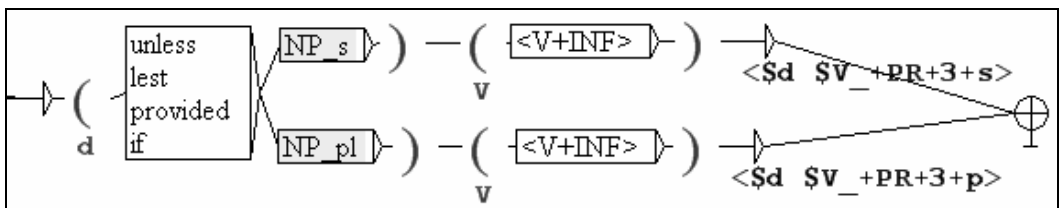

**Figure 7.** Graphe de reconnaissance du subjonctif.

–Formes avec soever

Les structures nominales suivantes sont remarquables: no <N0> of what <N1> + soever peut se traduire whatever the N1 of a N0. On trouve la structure « *at what great distance soever* » (Smith, 186), qui donnerait « *however great the distance is* » en anglais moderne, ou « *of what communion soever* », « *of whatever communion* » en anglais moderne, de même que la construction adjectivale similaire : *how <A> soever* se transpose en " *however adjectif* " : "How strict soever this Church is esteemed in admitting many degrees of marriage" (Rycaut). La Figure 8 présente ces graphes de reconnaissance et de réécriture de telles séquences en AC.

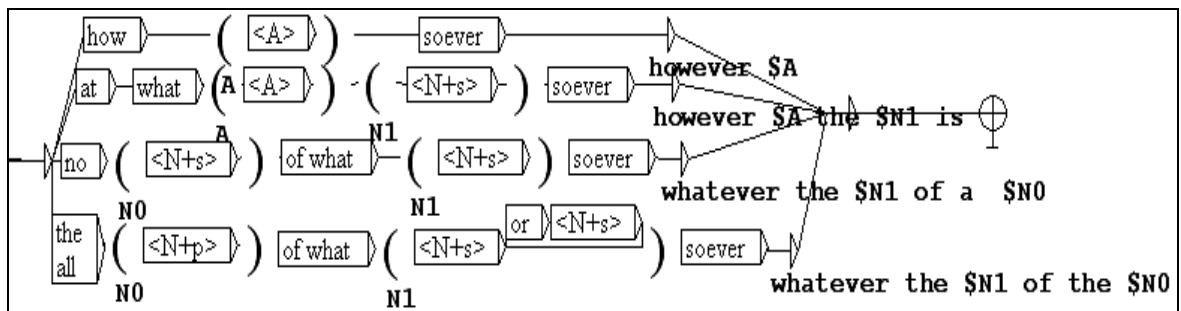

**Figure 8.** Graphe de reconnaissance des formes "soever".

–Formes déclarative, interrogative et négative.

À la forme déclarative du présent et du prétérit, à côté de la forme usuelle du verbe (avec la flexion "s" à la troisième personne du singulier), l'auxiliaire do(e) peut être utilisé avec l'infinitif: "They [the Greeks] are of diverse trades in cities, and in the country do till (=cultiver) the earth" (Sandys). "Mens minds did labour with fearefull expectations" (Sandys).

Nous pouvons traiter ces formes par le graphe présenté en Figure 9. Il reconnaît les séquences (do | doe | did | does | doth) [ADV] V. L'adverbe est optionnel. L'auxiliaire do est supprimé et le verbe est mis au temps voulu : lemme, prétérit, ou troisième personne du singulier du présent. Nous indiquons quelques exemples de résultats : doe still remain/<still remain>, doth believe them/<believes them> et did sit/<sat>.

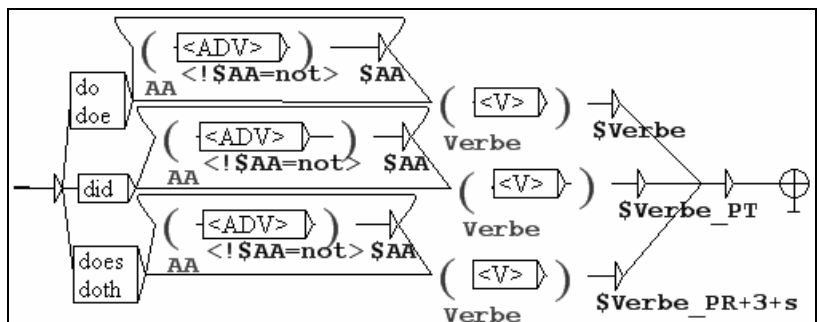

**Figure 9.** Graphe de transcription de la forme déclarative.

–Do est aussi utilisé à l'impératif: "do thou bless this oblation", "do thou pardon, as being our good and gracious God" (Smith). Un graphe permet de transcrire cette forme en AC en remplaçant do thou <V> par <V>.

–À la forme négative, à côté de l'auxiliaire "do" + not + verbe, on trouve aussi la structure verbe + not, comme dans "the women marry not till the age of 24" (Sandys), ou dans la phrase "their difference theologicall I enquired not" (Blount).

–À la forme interrogative, le verbe avec un sujet inversé est parfois utilisé. Dans la langue du XVII[e] siècle, les questions ne nécessitent pas toujours l'auxiliaire "do". Donnons quelques exemples tirés de deux pièces d'Aphra Behn : "What mean you by this language?" (*The Dutch Lover*, 1673). "How came she in?" (*Abdelazer or the Moor's Revenge*, 1676). Nous n'avons pas traité ces formes absentes de notre corpus.

### 3.4. Evaluation

Notre travail a été construit à partir d'un corpus de textes représentant 44300 mots environ. Nous présentons l'évaluation de notre traitement sur les textes des six auteurs sélectionnés. La Table 9 fournit un nombre de mots inconnus, ainsi que de leurs occurrences, pour chacun des six auteurs.

| Auteur | Année | Taille du corpus | Mots différents | Mots inconnus | occurrences des mots inconnus |
|---|---|---|---|---|---|
| **Blount** | 1636 | 1408 | 621 | 132 | 190 |
| **Sandys** | 1652 | 3910 | 1419 | 310 | 427 |
| **Rycaut** | 1679 | 12920 | 2270 | 345 | 656 |
| **Smith** | 1682 | 22640 | 4285 | 710 | 1266 |
| **Wheler** | 1682 | 2365 | 747 | 80 | 127 |
| **Ray** | 1693 | 1140 | 533 | 60 | 79 |

**Table 9.** Évaluation des mots inconnus dans le corpus traité.

Nous avons évalué les pourcentages des mots inconnus par rapport au nombre de mots différents et par rapport au corpus. Ces pourcentages sont en moyenne respectivement de 16,58 % et de 6,18 %. Nous avons appliqué nos dictionnaires et nos graphes à ces mots inconnus et nous évaluons les résultats dans la Table 10, en distinguant les mots étrangers, les noms propres, et les abréviations des mots du dix-septième siècle. Nous constatons que ce dernier pourcentage est plus grand pour les textes les plus anciens. Ces mots sont repérés soit par un dictionnaire *ad hoc*, soit par un graphe, les résultats que nous donnons ont été validés. Les formes ainsi reconnues par un graphe sont insérées dans les dictionnaires.

| Auteur | Mots inconnus | Langue étrangère | Noms propres | Abréviations | XVII[e] |
|---|---|---|---|---|---|
| **Blount** | 132 | 0,70% | 7,48% | 0,47% | 91,36% |
| **Sandys** | 310 | 3,43% | 12,15% | 0,66% | 82,56% |
| **Rycaut** | 345 | 21,25% | 14,81% | 0,87% | 60,98% |
| **Smith** | 710 | 13,69% | 18,75% | 0,60% | 63,91% |
| **Wheler** | 80 | 12,50% | 35,58% | 2,88% | 50,96% |
| **Ray** | 60 | 13,40% | 26,80% | 2,06% | 58,76% |

**Table 10.** Catégorie des mots inconnus.

Parmi les mots non reconnus nous devons faire un traitement particulier pour les formes autrefois écrites avec un tiret. En effet sur 128 formes x-y telles que 'mid-land' présentes dans les textes, nous avons identifié 43 formes aujourd'hui écrites xy comme 'midland'. Celles-ci sont également insérées dans le dictionnaire.

En ce qui concerne les erreurs, leur nombre est faible, ce qui s'explique par le fait que nos graphes et dictionnaires ont été construits pour traiter ce corpus. C'est sur le traitement d'autres corpus que pourrait se faire l'évaluation de notre travail.

Enfin certains problèmes sont liées au plurilinguisme du texte : in peut être du latin in nomine Patris, ou de l'italien Che fida in Grego, to peut être du grec : To katasphragisai to paidion. Les mots empruntés, généralement non reconnus, vont être traités, à tort, par les graphes, et vont augmenter le nombre d'erreurs. Il conviendrait d'encadrer les citations de marques permettant d'appliquer les dictionnaires adéquats.

Nous donnons ici un exemple extrait des textes traités, et indiquons en caractères gras les réécritures effectuées : … when Antipater, Perdiccas, Seleucus, Lysimachus, Antigonus, Ptolemey, and the rest of the successors of Alexander had shared his empire among them, they endeavoured as much as they could to plant his new-got **kingdomes/ kingdoms** with their **countrey-men/ countrymen**: whose posterity in part **remaineth/ remains** to this day, (though **vassaled/submitted** to the often changes of **forain/ foreign governours/governors: supplyed/ supplied** by the **extention/extension** of the latter Greek Empire; who yet retain wheresoever/**wherever** they live, their name, their religion, and particular language. A nation once so excellent, that their precepts and examples **doe/do** still remain as approved canons to direct the mind that **endevoureth/endeavours vertue/ virtue**. Admirable in arts and glorious in arms; famous for government, **affectors/lovers** of **freedome/ freedom**, every way noble: and to whom the rest of the world were reputed barbarians.

## 5. Conclusion
La partie automatisable de notre travail, même si elle apporte une aide précieuse, ne peut se concevoir que comme une étape suivie d'une validation. Les aspects grammaticaux ne sont actuellement pas tous traités. Si nous sommes en mesure de résoudre dans des proportions significatives l'insertion des apostrophes, et la réécriture de certaines formes archaïques, nous n'avons pas encore de moyen fiable de reconnaissance automatique des adjectifs postposés, l'utilisation d'adjectifs comme adverbes, et l'omission de prépositions. Le repérage des prépositions omises suppose un lexique-grammaire des verbes, le repérage de l'antécédent du relatif which pose problème, car il nécessite l'identification de traits humains et du genre dans les dictionnaires ce qui est un très gros travail. Enfin nous devons souligner que la combinaison des modifications que nous décrivons pose des problèmes autrement plus ardus que leur traitement séparé.

Dans ce travail la partie lexicographique est la plus aisée à mener à son terme. En élargissant le corpus et en créant une base de données consacrée aux récits de voyage anglais au XVII$^e$ siècle, nous voulons créer un dictionnaire qui inclut les archaïsmes et les mots étrangers et peut donner lieu à des éditions spécifiques. Le choix de traits sémantiques facilite l'établissement de glossaires spécialisés : glossaire religieux, ou de nature historique ou géographique. Lors de la recherche en ligne sur catalogue informatisé, le recensement des variantes orthographiques d'un même mot permettrait de retrouver des ouvrages en tapant un titre avec une orthographe moderne.

L'existence d'un dictionnaire de la langue anglaise du XVII$^e$ siècle serait d'un grand intérêt pédagogique. Il permettrait de recenser les principales particularités sémantiques et morphosyntaxiques de l'anglais du XVII$^e$ siècle et de proposer des équivalents en anglais moderne. Deviennent aisément lisibles et abordables par nos étudiants et par le grand public (anglophone ou non) des textes jusque-là oubliés ou ensevelis dans les Rare Books Rooms des bibliothèques anglaises ou accessibles seulement aux chercheurs dûment accrédités. Ajoutons que le corpus va s'élargir au fil du temps puisque nous travaillons désormais sur d'autres aires géographiques.

Nous espérons ainsi rendre les textes du XVII[e] siècle, et en particulier les récits de voyage, qui sont une source historique précieuse, plus accessibles aux étudiants non anglophones et non-spécialistes et leur éviter de recourir aux traductions de l'époque qui non seulement sont de belles infidèles mais omettent parfois de traduire des phrases ou des passages entiers du texte original.

## Références

---

[1] Oxford English Dictionary

[2] wives' (GB) ou (wives's) US. Notre objectif est de repérer le génitif, et pour en rendre possible le traitement automatique, nous appliquons la règle de *The Chicago Manual of Style* qui recommande l'addition du s, ayant bien conscience qu'il s'agit d'une norme américaine et non britannique.

[3] *Nous offrons à toi de ce qui est à toi…*